\documentclass{article}

\usepackage{times}
\usepackage{graphicx}
\usepackage{subfigure}
\usepackage{natbib}
\usepackage{algorithm}
\usepackage{algorithmic}
\usepackage{lipsum}
\usepackage{todonotes}
\usepackage[accepted]{icml2017}
\usepackage{booktabs}
\usepackage{makecell}
\usepackage{amsmath}

\newcommand{\icmltitlerunningtext}{CS 287 Final Project}
\icmltitlerunning{\icmltitlerunningtext}

\usepackage[hyphens]{url}
\usepackage{hyperref}
\hypersetup{colorlinks=true,breaklinks=true}

\begin{document}
\twocolumn[
\icmltitle{Pivot Language for Low-Resource Machine Translation}
\begin{icmlauthorlist}
  \icmlauthor{Abhimanyu Talwar}{}
\icmlauthor{Julien Laasri}{}
\end{icmlauthorlist}
\centerline{Harvard University}

\vskip 0.05in
\centerline{\small May 13, 2019}
\vskip 0.2in
]

\pagestyle{fancy} 
\fancyhead[C]{\nouppercase{\icmltitlerunningtext}}
\fancyhead[L]{} 
\fancyhead[R]{} 
\fancyfoot{} 

\begin{abstract}
Certain pairs of languages suffer from lack of a parallel corpus which is large in size and diverse in domain. One of the ways this is overcome is via use of a pivot language. In this paper we use Hindi as a pivot language to translate Nepali into English. We describe what makes Hindi a good candidate for the pivot. We discuss ways in which a pivot language can be used, and use two such approaches - the Transfer Method (fully supervised) and Backtranslation (semi-supervised) - to translate Nepali into English. Using the former, we are able to achieve a \textit{devtest} Set SacreBLEU score of 14.2, which improves the baseline fully supervised score reported by \cite{datasets} by 6.6 points. While we are slightly below the semi-supervised baseline score of 15.1, we discuss what may have caused this under-performance, and suggest scope for future work.

\end{abstract}

\section{Introduction}
\label{sec:introduction}


Machine Translation has seen a lot of improvements in the recent years mainly thanks to the introduction of Neural Machine Translation (NMT) systems that are trained on huge corpora of parallel data. On languages where such parallel data is available (e.g. French-English with 36M sentences), we now obtain human-level translation performance \cite{transformer}. Still, results are far behind when it comes to languages for which parallel corpora remain small (e.g. a few hundred thousand sentences). Part of the reason why research in this field is not progressing efficiently is the absence of high-quality evaluation sets for low-resource languages. To solve this problem, Facebook recently introduced two new datasets to evaluate low-resource machine translation on Nepali-English and Sinhala-English. In this paper, we decided to focus on the translation from Nepali to English but our work could be transcribed to other low-resource language pairs. The current best methods for these tasks include a fully supervised approach using only the available parallel corpus, a weakly supervised setting using cleaned data automatically crawled from the web \cite{zipporah} (e.g. Paracrawl\footnote{https://paracrawl.eu/}) and a semi-supervised technique that extends the parallel corpora using backtranslation \cite{backtranslation}. Our plan consists in the usage of Hindi as a pivot language for the Nepali-to-English translation task.

Our contributions include putting together and cleaning up datasets from three sources to create our Nepali-Hindi parallel corpus, which we believe is more diverse in domain and contains more well-formed sentences when compared with the available Nepali-English parallel corpus. We discuss various ways in which a pivot language can be used, and then use two approaches - Transfer Method (fully supervised) and Backtranslation (semi-supervised) to translate Nepali to English. Using the former, we are able to achieve a Test Set BLEU score of 14.2, which improves the baseline fully supervised score reported by \cite{datasets} by 6.6 points.

\section{Background}
\label{sec:background}

\begin{figure}
  \centering
  \includegraphics[width=1.0\linewidth]{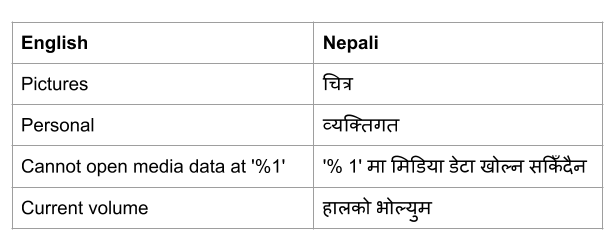}
  \caption{Examples of ill-formed sentences from the Nepali-English Parallel Corpus.}
  \label{fig:gnome-examples}
\end{figure}

The quality of Neural Machine Translation for certain language pairs is limited by the amount of parallel text available. For instance, the size of English-Nepali parallel corpus made available to us is 569k sentences, almost all of which comes from a single domain - the translation of Linux docs from English to Nepali. We note that a number of parallel sentence pairs in this corpus comprise of just a handful of words, and are not well-formed sentences. Fig. \ref{fig:gnome-examples} shows certain examples of ill-formed sentences from the Nepali-English parallel corpus. We believe such poor quality sentences may not help our translation models in learning word reordering and differences in morphological complexity, which we discuss in detail in Section \ref{sec:hindi}. 

Using a pivot language can help overcome these shortcomings by employing tricks such as backtranslation and synthetic data augmentation described in Section \ref{sec:related_work}. For our experiments, we have chosen Hindi as the pivot language. Our Nepali-Hindi dataset is more diverse in domain, and our Hindi-English dataset is larger in size and more diverse in domain, when compared with the Nepali-English parallel corpus available to us.

\subsection{Nepali-Hindi Parallel Corpus}
To create this dataset, we combined data sourced from OPUS\footnote{http://opus.nlpl.eu/} \cite{OPUS}, the Bible's massively parallel corpus\footnote{http://christos-c.com/bible/} \cite{bible}  and the Indian Language Technology Proliferation and Deployment Center (ILTPDC)\footnote{http://tdil-dc.in/index.php?lang=en}. Table \ref{fig:ne-hi-parallel} summarizes the statistics of our combined parallel corpus. We note that while some of the sentences sourced from Linux translations will be ill-formed (as in the case for Nepali-English shown in Table \ref{fig:gnome-examples}), the addition of ILTPDC data not only adds many well-formed sentence pairs to this dataset, but also broadens its domain by including sentences related to agriculture and entertainment.

\begin{table}[]
    \centering
    \begin{tabular}{lrrr}
        \toprule
         \textbf{Dataset} & \textbf{Train} & \textbf{Valid} & \textbf{Test}\\
         \midrule
         \makecell[l]{GNOME/KDE/Ubuntu} & 213K & 464 & 2344\\ 
         \midrule
         Bible & 30K & 65 & 330\\
         \midrule
         \makecell[l]{Agriculture and \\ Entertainment Domain} & 41K & 87 & 444\\
         \midrule
         \textbf{Total} & \textbf{284K} & \textbf{616} & \textbf{3118}\\
         \midrule
    \end{tabular}
    \caption{Number of sentences in our Nepali-Hindi Parallel Corpus}
    \label{fig:ne-hi-parallel}
\end{table}

\subsection{Hindi-English Parallel Corpus}
For Hindi-to-English, we have used the parallel corpus provided by \cite{iit-bombay}. Table \ref{fig:iit-bombay} shows the key statistics for this dataset, which is fairly large with nearly 1.5 million sentences. Further, as shown in Table \ref{fig:iit-domain}, this dataset is fairly diverse in terms of domain, compared with the Nepali-English parallel corpus.

\begin{table}
  \centering
  \begin{tabular}{lcrrr}
  \toprule
  & \textbf{Lang.} & \textbf{Train} & \textbf{Valid} & \textbf{Test} \\
  \midrule     
  \makecell[l]{No. of \\ Sentences} & & 1,492,827 & 520 & 2,507\\ 
  \midrule
  \makecell[l]{No. of \\ Tokens} & ENG & 20,667,259 & 10,656 & 57,803\\
  & HIN & 22,171,543 & 10,174 & 63,853\\
  \midrule
  \end{tabular}
  \caption{Statistics for the Training Set of IIT Bombay Hindi-English Parallel Corpus.}
  \label{fig:iit-bombay}
\end{table}

\begin{table}[]
    \centering
    \begin{tabular}{lr}
        \toprule
         \textbf{Data Source} & \textbf{Share of Corpus}  \\
         \midrule
         \makecell[l]{HindiEnCorp (News, Wikipedia, \\ Ted Talks etc.) + Ted Talks \\ + Wiki Headlines}
 & 23\% \\
        \midrule
         \makecell[l]{Parallel examples in dictionaries} & 20\% \\
         \midrule
         Gyaan-Nidhi (book translations) & 15\% \\
         \midrule
         Religious Texts & 13\% \\
         \midrule
         Other & 29\% \\
         \midrule
    \end{tabular}
    \caption{Domains comprising the IIT-Bombay Hindi-English Parallel Corpus}
    \label{fig:iit-domain}
\end{table}

\subsection{Monolingual data}
We also have access to a lot of monolingual data for each of these three languages: Nepali, Hindi and English. This type of data can easily be obtained by crawling the web (e.g. Wikipedia). The only monolingual data we ended up using was in Hindi to do backtranslation on the Nepali-Hindi language pair. Details of the data which was made available by IIT Bombay \cite{iit-bombay} can be found in table \ref{fig:hi-monolingual-data}.

\begin{table}
  \centering
  \begin{tabular}{lr}
  \toprule
  \textbf{Source} & \textbf{Sentences} \\
  \midrule     
  BBC-new & 18K\\ 
  \midrule
  BBC-old & 135K\\
  \midrule
  \makecell[l]{Hindi MonoCorp \\ \cite{hi-monocorp}} & 44.5M\\
  \midrule
  Health domain & 8K\\
  \midrule
  Tourism domain & 15K\\
  \midrule
  Wikipedia & 259K\\
  \midrule
  Judicial Domain & 153K\\
  \midrule
  \textbf{Total} & \textbf{45M}\\
  \midrule
  \end{tabular}
  \caption{Statistics for the IIT Bombay Hindi monolingual data.}
  \label{fig:hi-monolingual-data}
\end{table}

\section{Related Work}
\label{sec:related_work}

A few different approaches have been introduced to deal with low-resource language pairs. The most basic approach consists in training a simple model such as a Transformer \cite{transformer} on the available low-amount parallel data that we have access to. Unsupervised techniques based solely on monolingual data exist \cite{unsupervised} but do not perform well at all on these two new datasets \cite{datasets}. Better approaches generally try to augment the parallel corpus by introducing new noisy parallel data. 
One method adds to the existing parallel corpus new pair sentences automatically crawled from the web (see Paracrawl) and cleaned using the Zipporah method \cite{zipporah}. Unfortunately, as mentioned by the authors, this does not bring much improvement in the specific case of Nepali-to-English translation due to the extremely low availability of good parallel data for this pair language to train the cleaning method.

Another common trick to augment the parallel corpus is backtranslation \cite{backtranslation}. If our goal is to translate Nepali to English, we first train an English-to-Nepali translator using the available parallel corpus. We then backtranslate English monolingual data and add these artificial pair sentences to the parallel corpus to train a system that goes from Nepali to English. This process can be repeated multiple times to increasingly improve the translation results. There are two major benefits to this approach. The first one is that the target data is drawn from real English language so we are just improving the target language model using the available monolingual data. The second is that it augments the domain of the target language by allowing the model to train on out-of-domain data. In other words, if your parallel corpus is focused on one specific domain (e.g. the Ubuntu handbook) but you have diverse monolingual data coming from other sources (e.g. Wikipedia), the translation system can become better at translating sentences from this extended diverse domain.

All the techniques mentioned above have been tested on these new datasets and their scores were reported by \cite{datasets} and the best performing results can be found in table \ref{fig:mainres}.

A completely different approach consists in introducing a pivot language, in our case Hindi, to improve the translation systems with three possible methods. These methods were introduced for Statistical Machine Translation (SMT) systems but researchers have started using them with neural models as well when it is possible.

\subsection{Transfer method}
\label{sec:transfer}
The transfer method \cite{transfer} is the most straightforward approach one could think of when introducing a third pivot language to improve a translation system. In our case, it consists in training separately a Nepali-to-Hindi and a Hindi-to-English model. Given a source sentence $s$ in Nepali, we produce $n$ pivot sequences $p_1, \ldots, p_n$ in Hindi using the first model. Then, for each pivot sentence $p_i$, we produce $m$ target English sentences $t_{i, 1}, \ldots, t_{i, m}$. These $n \times m$ candidates are then scored \cite{triangulation} to find the best translation possible of the Nepali source sentence $s$ to an English sentence. Using $n=m=1$ allows us to get rid of that scoring procedure.

\subsection{Synthetic method}
\label{sec:synthetic}
The synthetic method \cite{synthetic} can be divided into two approaches.

\begin{itemize}
    \item \textbf{Synthetic target:} we create a Nepali-SyntheticEnglish corpus by using a Hindi-to-English translation system on Nepali-Hindi parallel data.
    \item \textbf{Synthetic source:} we create a SyntheticNepali-English corpus by using a Hindi-to-Nepali translation system on Hindi-English parallel data.
\end{itemize}

We can then add the synthetic data coming from one of these two approaches to the true available parallel data in order to train the actual translation system we're interested in.

In our case, we would expect the synthetic source approach to work better than the synthetic target one for two reasons:
\begin{itemize}
    \item Each of these two approaches requires a good corresponding translation system and a good amount of corresponding parallel data. We expect to have a good Hindi-to-Nepali translation system thanks to the relatedness of the two languages and we have access to significantly more high-quality parallel data of Hindi-English than Nepali-Hindi. See Section \ref{sec:background} for more details on the datasets.
    \item The synthetic source preserves real sentences in the target language (English here) and would thus help improve the language model of the target, similar to what was happening with backtranslation.
\end{itemize}

\subsection{Triangulation method}

This method can be particularly useful when dealing with Phrase-Based SMT systems. As described in \cite{triangulation}, we can build a phrase table for Nepali-English based on two pivot phrase tables corresponding to Nepali-Hindi and Hindi-English. To do so, given a phrase $\bar{n}$ in Nepali and a candidate phrase $\bar{e}$ in English, the probabilities of the Nepali-English phrase table can be estimated as follows:
$$
    \phi(\bar{e}|\bar{n}) = \sum_{\bar{h} \in T_{EH} \cap T_{HN}} \phi(\bar{e}|\bar{h})\phi(\bar{h}|\bar{n})
$$

where $T_{EH} \cap T_{HN}$ are the English phrases that are present in both Nepali-Hindi and Hindi-English (finite) phrase tables.

With neural models, we directly predict full sentences. As a result, in the NMT world, the computation above would become: 
$$
    \phi(e|n) = \sum_{h \in \text{all hindi sentences}} \phi(e|h) \phi(h|n)
$$
where $n$ is a given input sentence in Nepali, and $e$ is a candidate English sentence. This computation is intractable so we cannot directly apply this method to a NMT system.

\section{Models}

We created two models using the transfer method for this paper. More details on the training of these models can be found in Section \ref{sec:training}.

\subsection{Fully-Supervised}

Our fully-supervised model corresponds to the usage of the transfer method with $n=m=1$ on two Transformer models independently trained on our available corpora of Nepali-Hindi and Hindi-English.

\subsection{Semi-Supervised}

Our semi-supervised model is also a transfer method with $n=m=1$. It uses the same Hindi-to-English Transformer as our fully-supervised model. The difference is that is uses one iteration of backtranslation for the training of the Nepali-to-English Transformer. As shown in Section \ref{sec:background}, we already have a significant amount of diverse parallel sentences for the Hindi-English language pair. On the contrary, our Nepali-Hindi parallel corpus is quite limited and not diverse at all. We thus randomly subsampled 300K sentences from the monolingual Hindi data made available by IIT Bombay. As shown by table \ref{fig:hi-monolingual-data} and \cite{hi-monocorp}, this data is large and diverse. The idea of using backtranslation with this monolingual data on the Nepali-Hindi language pair was thus to improve the diversity of our parallel corpus while making it larger.





\section{Hindi as a Pivot Language}
\label{sec:hindi}
We have chosen Hindi as the pivot language for translation between Nepali and English. Our choice is driven by the high degree of "relatedness" between Nepali and Hindi, and the availability of large Hindi datasets  (mono-lingual as well as parallel with English/Nepali). In  \cite{pivot-relatedness}, the authors have shown that language translation benefits from certain dimensions of relatedness between the source and target. We discuss how Hindi fares along these dimensions below.

\subsection{Historical Relatedness}
Nepal shares part of its border with northern India and by some estimates, nearly 40\% of Nepal's population also speaks Hindi. Both languages are written in the same script of Devanagari, and if someone can read one language then they can also read the other. They also have a high degree of lexical similarity, i.e. they share a large part of their vocabularies \cite{lexical-similarity}. 
\subsection{Amount of Reordering}
\begin{figure}
  \centering
  \includegraphics[width=1.0\linewidth]{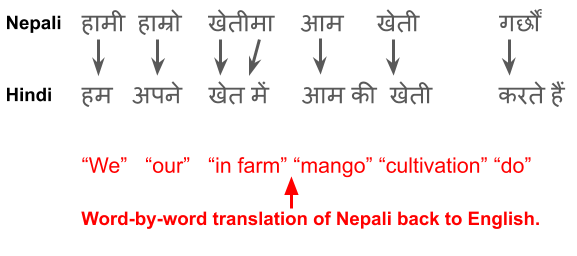}
  \caption{The amount of word reordering required is minimal when translating between Hindi and Nepali but is significant when translating between Nepali and English.}
  \label{fig:reordering}
\end{figure}

This pertains to how much reordering of corresponding words is required when we translate a sentence from one language to another. Fig. \ref{fig:reordering} shows Nepali and Hindi translations of the English sentence "We cultivate mangoes in our farm", in the top and next rows respectively. We observe that the order of corresponding words is the same between Nepali and Hindi (as shown by the arrows). However, if we translate the Nepali sentence to English, word-by-word, we end up with "We our in farm mango cultivation do". This indicates that a high amount of reordering is required when translating to English. We note that this is not an anecdotal occurrence and is in fact a feature of the languages themselves. English follows a Subject-Verb-Object structure while Hindi and Nepali follow a Subject-Object-Verb structure.

\subsection{Morphological Complexity}
\begin{figure}
  \centering
  \includegraphics[width=1.0\linewidth]{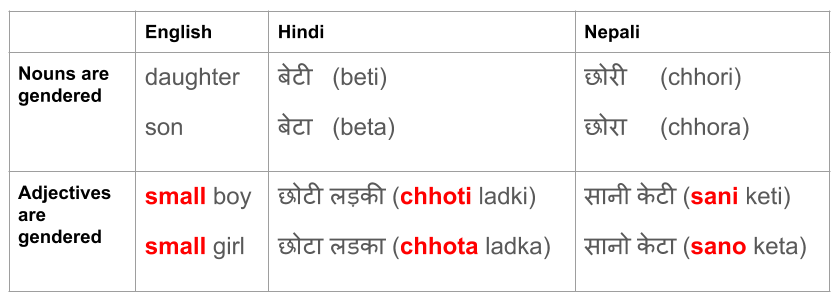}
  \caption{Nepali and Hindi share several morphological features and are different from English.}
  \label{fig:morphology}
\end{figure}

Morphology pertains to how words in a language are formed from simpler primitives (e.g. "cats" is formed from the singular "cat"). Here again, Hindi and Nepali are fairly similar to each other while being different from English. For instance, nouns and adjectives are gendered in Nepali and Hindi while that is not the case for English. Fig. \ref{fig:morphology} shows that when we want to say "small boy" or "small girl", we use the same adjective "small", however in each of Hindi and Nepali, the corresponding adjective for "small" is different when we are talking about the female gender compared to when we talk about the male gender. Another example in Fig. \ref{fig:morphology} is about gendered nouns - in English we have completely different words to refer to the male child ("son") and female child ("daughter") whereas in both Hindi and Nepali, the corresponding words are quite similar - they share the same root word and only differ in the suffix to that root. 

\section{Training and Inference}
\label{sec:training}

\subsection{Preprocessing}
We preprocess our data in a manner similar to that used by \cite{datasets}. Specifically, for each of the language pairs, Nepali-Hindi and Hindi-English, we learn a joint Byte Pair Encoding (BPE) vocabulary of 5,000 symbols using the SentencePiece library\footnote{https://github.com/google/sentencepiece}. The BPE vocabulary is learnt over raw English sentences and tokenized Hindi/Nepali sentences. We have used scripts included within Facebook AI Research's (FAIR) Flores package\footnote{https://github.com/facebookresearch/flores} for the aforementioned preprocessing steps.

\subsection{Model Architecture and Hyperparameters}
For all our translation models, we have used the Transformer architecture proposed by \cite{transformer}, with hyperparameters as used by \cite{datasets}. Specifically, this Transformer has 5 encoder and 5 decoder layers, with the number of attention heads being 2. Further, the dimensions of token embeddings, and hidden states are 512 and 2048 respectively. We have used the implementation provided by Facebook AI Research's FairSeq package\footnote{https://github.com/pytorch/fairseq}. For each pair of languages, the model is trained by minimizing Label Smoothed Cross Entropy loss, with a smoothing parameter of 0.2. As described in \cite{label-smoothing}, this loss criterion does not use the ground truth labels for training but instead uses a weighted average of the ground truth label and a distribution which is uniform over the labels, with the weight controlled by the smoothing parameter. The models were trained using the Adam optimizer with betas of $(0.9, 0.98)$. The regularization scheme involved weight-decay of $1e-4$ and FairSeq dropout parameters of: Dropout (0.4), Attention-Dropout (0.2) and ReLU-Dropout (0.2). We trained all our models for 100 epochs except the Hindi-to-English model which were able to train for only 55 epochs due to time constraints. 

\subsection{Inference}
At test time, we have used the Transfer method described in Section \ref{sec:transfer}, with the value of each of the parameters $m$ and $n$ being 1.




\section{Results}


The results we get can be found in table \ref{fig:mainres}. Let's note that the task on the \textit{dev} set seems significantly more difficult than on the \textit{devtest} set. The baseline scores on the \textit{dev} set were not reported by \cite{datasets}.

\begin{table*}
  \centering
  \begin{tabular}{lrrrr}
  \toprule
      \textbf{Model} &  \textbf{Ne $\rightarrow$ Hi} Test & \textbf{Hi $\rightarrow$ En} Test & \textbf{Ne $\rightarrow$ En} \textit{dev} & \textbf{Ne $\rightarrow$ En} \textit{devtest}\\
  \midrule
      BASELINE (Fully-Supervised) &  - & - & - & 7.6\\
      BASELINE (Semi-Supervised)  &  - & - & - & 15.1\\
      OUR-Transfer Method (Fully-Supervised) & 47.2 & 16.8 & 11.3 & 14.2\\
      OUR-Transfer Method (Semi-Supervised) & 43.1 & 16.8 & 8.5 & 10.7\\
  \end{tabular}
  \caption{Results of the transfer method and comparison with the baselines. Baseline results are directly reported from \cite{datasets}. Transfer (Fully-Supervised) correspond to the transfer method with $n=m=1$. Transfer (Semi-Supervised) uses one iteration of backtranslation in the Nepali-Hindi translation procedure. We report tokenized BLEU \cite{bleu} for Ne$\rightarrow$ Hi and detokenized SacreBLEU for \{Ne, Hi\} $\rightarrow$ En \cite{sacrebleu}.}
  \label{fig:mainres}
\end{table*}

\section{Discussion}


We can see that our fully-supervised transfer model greatly improves the results in comparison with the fully-supervised baseline of \cite{datasets}. However, backtranslation seemed to be harmful to the Nepali-Hindi translation and the full model overall. What makes Nepali-Hindi translation so easy is the close similarity between the languages. Backtranslation introduces noise in the source which in turn decreases this close similarity and we believe that is what was harmful to the translation. One question that remains open is whether or not adding backtranslation on the Hindi-English part of the transfer may help our full model. We believe that the parallel data that we have is already diverse and large enough so that backtranslation would not be helpful, but the process could still potentially improve our target language model for English. We did not take the time to try this approach as backtranslation is extremely time-consuming, especially when it comes to parallel corpora which are already quite large.


\section{Conclusion}


In the end, we managed to beat the fully-supervised baseline using our fully-supervised transfer method by 6.6 points. We are still slightly under the semi-supervised baseline. We believe that our score can be improved simply by letting our Hindi-English Transformer train for a few more days until we reach convergence, probably at around 100 epochs. Another future work that we've started working on is the implementation of the Synthetic method, and more specifically its Synthetic Source approach. As discussed in Section \ref{sec:synthetic}, we believe we have the necessary data to get good results with this technique. 




\bibliography{example}
\bibliographystyle{icml2017}

\end{document}